\documentclass{vgtc}                          

\graphicspath{{figures/}{pictures/}{images/}{./}} 

\usepackage{times}                     

\usepackage{tabu}                      
\usepackage{booktabs}                  
\usepackage{lipsum}                    
\usepackage{mwe}                       
\usepackage{adjustbox}
\usepackage{multirow}
\usepackage{array}
\usepackage{float}
\usepackage{makecell}
\usepackage{amsmath}

\usepackage{mathptmx}                  

\onlineid{0}

\vgtccategory{Research}

\vgtcinsertpkg

\title{SciVisAgentSkills: Design and Evaluation of Agent Skills for\\ Scientific Data Analysis and Visualization}

\author{
Kuangshi Ai\thanks{e-mail: kai@nd.edu}\\
    \scriptsize Univ. Notre Dame %
\and Haichao Miao\thanks{e-mail: miao1@llnl.gov}\\
    \scriptsize LLNL%
\and Kaiyuan Tang\thanks{e-mail: ktang2@nd.edu}\\
    \scriptsize Univ. Notre Dame %
\and Shusen Liu\thanks{e-mail: liu42@llnl.gov}\\
    \scriptsize LLNL %
\and Chaoli Wang\thanks{e-mail: chaoli.wang@nd.edu}\\
    \scriptsize Univ. Notre Dame
}

\abstract{
Recent advances in agentic visualization have enabled the translation of natural language into executable scientific visualization (SciVis) workflows. While general-purpose coding agents show strong capabilities, they often lack the tool-specific expertise required for SciVis tasks. In this work, we present \textit{SciVisAgentSkills}, a collection of reusable agent skills that augment coding agents for scientific data analysis and visualization by encoding environment assumptions, tool usage patterns, and domain heuristics across scientific tools such as ParaView, napari, VMD, and TTK. We evaluate these skills on Codex and Claude Code using SciVisAgentBench, a benchmark of 108 expert-designed multi-step tasks. Results show that agent skills improve mean task scores across the evaluated suites, with token-efficiency benefits that depend on the agent harness and tool setting. These findings highlight the importance of structured procedural knowledge for enabling reliable, long-horizon SciVis workflows, while also showing that skills should be studied alongside the execution harness that loads and applies them. The skills are available at \url{https://github.com/KuangshiAi/SciVisAgentSkills}.
} 


\begin{document}


\maketitle

\section{Introduction}

Multimodal large language models (MLLMs) are increasingly being used to power \emph{agentic visualization} systems, where (semi-)autonomous agents convert natural-language requests into concrete visualization actions~\cite{dhanoa2025agentic, liu2024ava, peterka2025chatvis, sun2026sasav}.
Scientific visualization (SciVis) transforms complex scientific data, such as volumes, meshes, flow fields, and molecular structures, into interpretable visual representations for analysis and discovery. As simulations and instruments generate increasingly complex data, SciVis has become essential for exploring, validating, and communicating findings across disciplines such as physics, biology, climate science, and materials science.
In SciVis, tasks often require long, multi-step workflows combining domain knowledge and tool interaction. At the same time, the emergence of the model context protocol (MCP)~\cite{anthropic2024mcp} has made complex visualization tools more accessible by enabling agents to interact with software via structured interfaces. Recent MCP agents~\cite{liu2025paraview, llnl2025bioimageagent, egtai2025gmxvmdmcp, gorski2026topopilot} and agentic systems~\cite{ai2025nli4volvis, biswas2025vizgenie, tam2025infera} have demonstrated the feasibility of long-horizon SciVis workflows.

Alongside these domain-specific systems, general-purpose coding agents are becoming increasingly capable. These agents can already execute commands, inspect files, write code, and iteratively refine outputs. Rather than building a custom agent for every tool or task, one can start from a strong general-purpose harness and adapt it to SciVis workflows through procedural guidance. This motivates our central question: how can general-purpose agents be equipped with the domain knowledge needed for effective and efficient scientific data analysis and visualization?

One promising answer is the use of \emph{agent skills}. An agent skill is a structured package of instructions, code templates, references, and verification logic that augments agent behavior at inference time without changing model parameters~\cite{anthropic2025agentskillsblog}. Agent skills capture procedural knowledge such as recommended workflows, environment-specific conventions, software usage patterns, and domain heuristics. In this sense, agent skills serve as a reusable layer between a base model and a specialized application domain: foundation models provide general capabilities, agent harnesses manage context and tool execution, and agent skills inject task- and domain-specific expertise.

The broader agent skill ecosystem has expanded rapidly, with community repositories now hosting thousands of user-contributed agent skills for software engineering, analysis, and enterprise workflows. Prior work suggests that general-purpose coding agents such as Claude Code~\cite{anthropic2025claudecode}, Codex~\cite{openai2025codex}, and Gemini CLI~\cite{google2025geminicli} are already competitive on long-horizon visualization tasks, and that agent skills can further improve both success rate and efficiency~\cite{ai2026scivisagentbench}. Recent work also compares interaction paradigms for SciVis agents and shows tradeoffs among structured tool use, CLI- or GUI-based general interaction, and persistent memory~\cite{vonderhorst2026exploring}. Despite this momentum, SciVis remains underexplored from the perspective of reusable agent skills. There are still a few published skills tailored to core SciVis workflows, and limited evidence on how portable skills behave across different tools, harnesses, and task types.

To address this gap, we present \textit{SciVisAgentSkills}, a collection of agent skills designed specifically for scientific data analysis and visualization. Our agent skills target core operations across several representative SciVis environments, including basic operations such as volume rendering, isosurface extraction, flow visualization, and scientific plotting in ParaView~\cite{Ahrens2005ParaView}, molecular visualization in VMD~\cite{VMD1996}, bioimage visualization in napari~\cite{napari2019}, and topology visualization in TTK~\cite{ttk2018}. We build on SciVisAgentBench, which introduced a napari skill case study and baseline measurements for coding agents~\cite{ai2026scivisagentbench}. We release a broader multi-tool skill collection, document the common design protocol used to produce it, and evaluate its behavior across agents, task suites, and token budgets. Finally, we evaluate Claude Code and Codex on SciVisAgentBench~\cite{ai2026scivisagentbench} with and without these agent skills. Our results show higher mean benchmark scores with skills across the evaluated suites, while token usage varies by agent harness and tool. More broadly, we position this paper as both an empirical study of SciVis skills in current coding-agent harnesses and a call for collaboration on building a reusable skill ecosystem for scientific data analysis and visualization.

\begin{figure*}[htb]
\centering
\includegraphics[width=\linewidth]{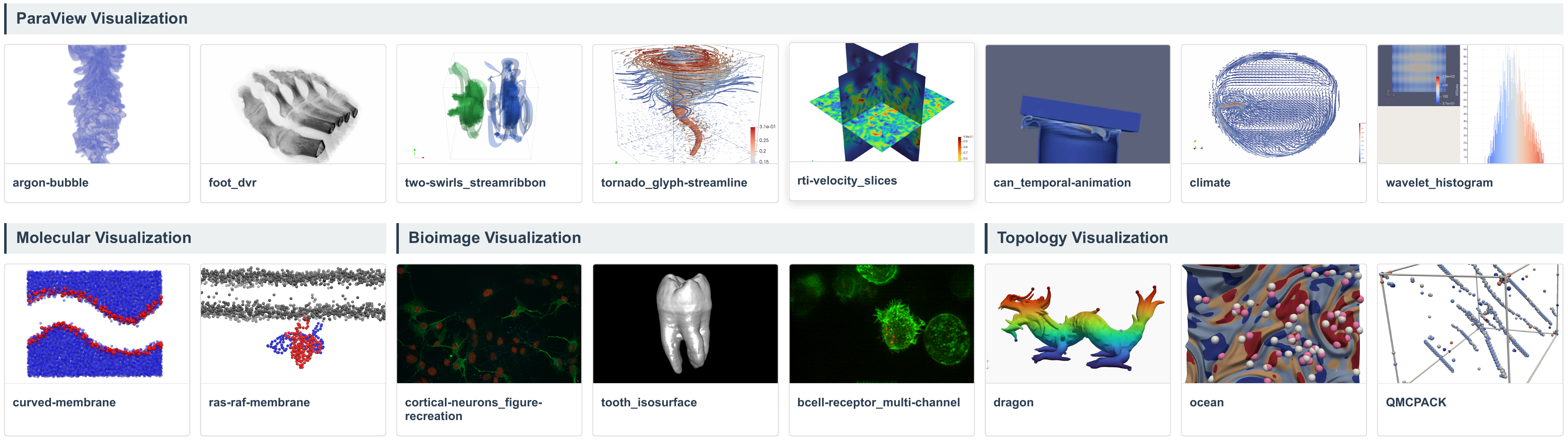}
\vspace{-0.25in}
\caption{Representative scientific data analysis and visualization tasks from SciVisAgentBench.}
\label{fig:example_cases}
\end{figure*}

\section{Related Work}

{\bf Agent evaluation and agent skills.} 
General-agent benchmarks such as AgentBench~\cite{liu2023agentbench}, GAIA~\cite{mialon2023gaia}, and $\tau$-bench~\cite{yao2024tau} evaluate multi-turn reasoning and tool use, but treat visualization as a generic task and do not capture multi-step SciVis workflows or domain-specific interpretation. Recent work on agent skills~\cite{jiang2026sok,xu2026agent} instead models agent behavior as reusable procedural modules loaded at inference time. SkillsBench~\cite{li2026skillsbench} shows that curated skills improve performance, while self-generated skills are often unreliable; prior analysis~\cite{ling2026agent} further shows that existing skill ecosystems remain concentrated in software engineering.

{\bf Agentic workflows for visualization.}
Dhanoa et al.~\cite{dhanoa2025agentic} frame this space as \textit{agentic visualization}, balancing automation and analyst control. Existing systems include conversational and collaborative analysis tools (VOICE~\cite{jia2025voice}, IntuiTF~\cite{wang2025intuitf}, CoDA~\cite{chen2025coda}), tool-centered SciVis assistants (AVA~\cite{liu2024ava}, ChatVis~\cite{peterka2025chatvis}, ParaView-MCP~\cite{liu2025paraview}), and more autonomous or multi-agent workflows (VizGenie~\cite{biswas2025vizgenie}, NLI4VolVis~\cite{ai2025nli4volvis}, TexGS-VolVis~\cite{tang2025texgs}, InferA~\cite{tam2025infera}, SASAV~\cite{sun2026sasav}, VIS co-scientist~\cite{miao2026toward}). Vonderhorst et al.~\cite{vonderhorst2026exploring} further compare interaction paradigms for SciVis agents, including structured tool use, CLI/GUI interaction, and persistent memory. While these systems demonstrate the promise of agentic SciVis, most are tailored to specific tools, interfaces, or applications. We instead focus on portable agent skills that improve general-purpose coding agents across multiple SciVis tools.

{\bf Agentic visualization evaluation.}
Existing evaluations focus on charts, plotting code, or human literacy rather than long-horizon SciVis agents. VisEval~\cite{chen2024viseval} and Drawing Pandas~\cite{galimzyanov2025drawing} study chart understanding and code generation, while LIDA~\cite{dibia2023lida} introduces visualization-specific metrics and SVLAT~\cite{do2026svlat} measures human SciVis literacy. Ai et al.~\cite{Ai-GenAI25} call for systematic evaluation of agentic SciVis. In response, NL2SciVis~\cite{mathai2026nl2scivis} introduces a benchmark for evaluating natural-language-driven SciVis in ParaView through atomic operations with deterministic validation. SciVisAgentBench~\cite{ai2026scivisagentbench} further extends this direction to multi-step tasks spanning data analysis, tool use, and end-to-end SciVis workflows, and serves as the testbed for our study of agent skills.

\begin{figure*}[t]
\centering
\includegraphics[width=\linewidth]{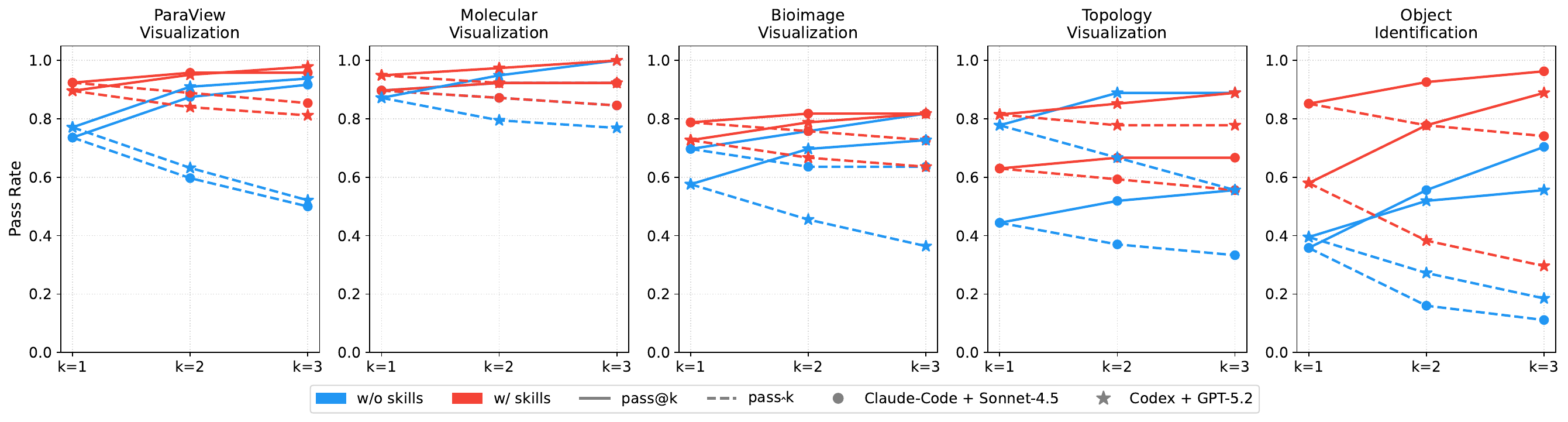}
\vspace{-0.25in}
\caption{pass@$\{1,2,3\}$ (i.e., success in at least one of the first $k$ trials) and pass\string^${\{1,2,3\}}$ (i.e., success in all $k$ trials) results of coding agents with and without SciVisAgentSkills across five SciVisAgentBench task suites.}
\label{fig:pass_k_combined}
\end{figure*}

\section{SciVisAgentSkills: Design and Evaluation}


{\bf Agent skills for SciVis tools.}
To enable general-purpose coding agents to operate effectively in SciVis environments, we design a set of domain-specific agent skills tailored to four representative tools: ParaView~\cite{Ahrens2005ParaView}, napari~\cite{napari2019}, VMD~\cite{VMD1996}, and TTK~\cite{ttk2018}. These tools collectively cover core SciVis workflows, bioimage analysis, molecular visualization, and topology visualization.

Our design is motivated by several recurring failure modes observed when general-purpose coding agents interact with these tools. Even in well-configured environments, agents often spend multiple turns probing libraries and execution settings, leading to unnecessary overhead. Due to incomplete grounding in tool-specific documentation, agents may misuse APIs or follow incorrect usage patterns. We also observe errors in output generation, such as capturing the entire napari GUI rather than the visualization viewport. These issues highlight the need for structured, domain-aware guidance beyond generic coding ability.

To address these challenges, we construct agent skills that encode environment assumptions, usage conventions, and best practices for each tool. We fix software versions and specify the execution environment, including dependencies and runtime constraints. This reduces ambiguity and redundant exploration. We then distill official documentation into structured usage patterns that guide agent behavior. To further ground execution, we incorporate representative code snippets and function usage patterns from existing SciVis agents, including ParaView-MCP~\cite{liu2025paraview}, BioImage-Agent~\cite{llnl2025bioimageagent}, GMX-VMD-MCP~\cite{egtai2025gmxvmdmcp}, and TopoPilot~\cite{gorski2026topopilot}. These examples provide references for common workflows and reduce trial-and-error during task execution. We also incorporate constraints from empirical observations, such as enforcing headless rendering and capturing outputs from the visualization viewport.

We therefore define an agent skill operationally as a self-contained, version-pinned procedural module for one SciVis tool, constructed through a unified process that combines environment specification, documentation alignment, executable exemplars, and failure-aware refinements. Following this process, we develop four agent skills corresponding to ParaView, napari, VMD, and TTK. Each skill encapsulates procedural knowledge for interacting with the target tool, providing a reusable layer that enables general-purpose coding agents to execute SciVis tasks more reliably and efficiently.

{\bf Skill format and construction.}
All four skills follow the agent skill format~\cite{anthropic2025agentskillsblog}: YAML frontmatter for discovery, followed by Markdown guidance that includes usage rules, script templates, API summaries, and troubleshooting notes. The ParaView skill additionally includes separate reference files because its API surface is larger. In the context of coding agents, this structure supports progressive disclosure, where the skill metadata is checked first, and the body and references are loaded only when relevant.

The skills were manually authored by visualization researchers and refined through several rounds of observing agents on representative SciVis workflows. Refinements targeted recurring tool-use failures, such as incorrect headless rendering and API misuse. To avoid benchmark leakage, the skills contain only tool-general procedural knowledge, including environment assumptions, documented API patterns, and reusable examples. They do not include benchmark-specific solutions, expected outputs, hidden labels, or evaluation rubrics. Exemplars are adapted from official documentation and public SciVis agent code~\cite{liu2025paraview, llnl2025bioimageagent, egtai2025gmxvmdmcp, gorski2026topopilot}. Before evaluation, the authors reviewed and tested each skill for correctness and safe headless execution in sandboxes.

{\bf Evaluating agent skills on SciVisAgentBench.}
We evaluate the effectiveness of our agent skills by comparing Codex and Claude Code with and without skills on SciVisAgentBench~\cite{ai2026scivisagentbench}, a benchmark designed for realistic, multi-step SciVis workflows. Unlike prior benchmarks that focus on short-horizon or 2D visualization tasks, SciVisAgentBench comprises 108 expert-crafted cases spanning diverse domains, data types, and visualization operations, with an emphasis on outcome-centric evaluation.

The benchmark includes five task suites. The ParaView visualization suite focuses on core operations, including volume rendering, isosurface extraction, flow visualization, and data analysis. Additional suites target molecular visualization (VMD), bioimage visualization (napari), and topology visualization (TTK). An object identification suite further evaluates whether agents can construct visualizations from anonymized volumetric data and infer the underlying object category from visual evidence, using the benchmark's hidden labels and case-specific rubrics for outcome evaluation. Representative examples are shown in Figure~\ref{fig:example_cases}.

SciVisAgentBench provides a suitable testbed for agent skills because its tasks are grounded in real tools and require coordinated, multi-step workflows, making performance sensitive to procedural knowledge. Its outcome-centric evaluation framework, combining multimodal LLM judges and deterministic metrics, enables reliable comparison of execution quality and efficiency without relying on unstable trajectory-level analysis.

Each task suite aligns naturally with one of our SciVisAgentSkills: the ParaView skill supports ParaView visualization and object identification; the napari skill targets bioimage visualization; the VMD skill supports molecular visualization; and the TTK skill addresses topology visualization. This alignment enables a direct with-versus-without comparison of domain-specific guidance under the same benchmark protocol.

\begin{table}[t]
\centering
\small
\caption{Benchmark performance across five task suites of SciVisAgentBench using {\bf Claude-Opus-4.6} as the LLM judge. For each agent+model setting, we report the overall score and completion rate as mean$\pm$std across three repeated trials.}
\label{tab:main_results}
\begin{adjustbox}{width=\columnwidth}
\begin{tabular}{clcc}
\toprule
Task Suite & Setting & Overall Score $\uparrow$ & Completion Rate $\uparrow$ \\
\midrule

\multirow{4}{*}{\shortstack{ParaView\\Visualization}}
& Claude-Code+Sonnet-4.5 (w/o skills) & 62.57$\pm$0.51 & 97.92$\pm$2.08 \\
& Codex+GPT-5.2 (w/o skills)          & 60.17$\pm$1.43 & 95.14$\pm$2.41 \\
& Claude-Code+Sonnet-4.5 (w/ skills)  & \textbf{73.93$\pm$2.56} & 95.83$\pm$3.61 \\
& Codex+GPT-5.2 (w/ skills)           & 66.67$\pm$1.10 & \textbf{98.61$\pm$1.20} \\

\midrule

\multirow{4}{*}{\shortstack{Molecular\\Visualization}}
& Claude-Code+Sonnet-4.5 (w/o skills) & 61.47$\pm$6.78 & 94.87$\pm$4.44 \\
& Codex+GPT-5.2 (w/o skills)          & 62.30$\pm$6.32 & 94.87$\pm$8.88 \\
& Claude-Code+Sonnet-4.5 (w/ skills)  & 64.33$\pm$4.88 & 92.31$\pm$0.00 \\
& Codex+GPT-5.2 (w/ skills)           & \textbf{73.13$\pm$3.25} & \textbf{100.00$\pm$0.00} \\

\midrule

\multirow{4}{*}{\shortstack{Bioimage\\Visualization}}
& Claude-Code+Sonnet-4.5 (w/o skills) & 52.83$\pm$9.80 & 90.91$\pm$9.09 \\
& Codex+GPT-5.2 (w/o skills)          & 41.90$\pm$4.69 & 75.76$\pm$10.50 \\
& Claude-Code+Sonnet-4.5 (w/ skills)  & \textbf{58.97$\pm$4.56} & 90.91$\pm$9.09 \\
& Codex+GPT-5.2 (w/ skills)           & 55.00$\pm$2.59 & \textbf{93.94$\pm$5.25} \\

\midrule

\multirow{4}{*}{\shortstack{Topology\\Visualization}}
& Claude-Code+Sonnet-4.5 (w/o skills) & 45.23$\pm$8.81 & 59.26$\pm$6.42 \\
& Codex+GPT-5.2 (w/o skills)          & 76.43$\pm$10.06 & \textbf{85.19$\pm$6.42} \\
& Claude-Code+Sonnet-4.5 (w/ skills)  & 73.63$\pm$3.85 & 70.37$\pm$6.42 \\
& Codex+GPT-5.2 (w/ skills)           & \textbf{83.73$\pm$4.75} & 81.48$\pm$6.42 \\

\midrule

\multirow{4}{*}{\shortstack{Object\\Identification}}
& Claude-Code+Sonnet-4.5 (w/o skills) & 41.50$\pm$3.55 & 83.95$\pm$9.32 \\
& Codex+GPT-5.2 (w/o skills)          & 43.33$\pm$5.28 & 92.59$\pm$9.80 \\
& Claude-Code+Sonnet-4.5 (w/ skills)  & \textbf{69.13$\pm$1.56} & \textbf{92.59$\pm$3.70} \\
& Codex+GPT-5.2 (w/ skills)           & 47.77$\pm$0.25 & 80.25$\pm$5.66 \\

\bottomrule
\end{tabular}
\end{adjustbox}
\end{table}

\begin{table}[t]
\centering
\small
\caption{Image-based evaluation metrics on ParaView visualization tasks. Values are reported as mean$\pm$std across three repeated trials. 
}
\label{tab:paraview_image_metrics}
\begin{adjustbox}{width=\columnwidth}
\begin{tabular}{lccc}
\toprule
Setting & $\mathrm{PSNR}_{\text{scaled}} \uparrow$ & $\mathrm{SSIM}_{\text{scaled}} \uparrow$ & $\mathrm{LPIPS}_{\text{scaled}} \downarrow$ \\
\midrule
Claude-Code+Sonnet-4.5 (w/o skills) & 20.99$\pm$0.68 & 0.92$\pm$0.02 & 0.10$\pm$0.02 \\
Codex+GPT-5.2 (w/o skills) & 21.27$\pm$1.02 & 0.92$\pm$0.02 & 0.10$\pm$0.03 \\
Claude-Code+Sonnet-4.5 (w/ skills) & \textbf{22.08$\pm$0.73} & 0.91$\pm$0.04 & 0.10$\pm$0.03 \\
Codex+GPT-5.2 (w/ skills) & 21.76$\pm$1.17 & \textbf{0.93$\pm$0.02} & \textbf{0.09$\pm$0.02} \\
\bottomrule
\end{tabular}
\end{adjustbox}
\end{table}

\begin{table}[t]
\centering
\small
\caption{Token usage across all five task suites of SciVisAgentBench. Input and output tokens are reported as mean$\pm$std across three repeated trials. Token counts are shown using K (thousands) and M (millions) for readability. Cached tokens are counted as regular input tokens for consistent accounting and comparison across settings. 
}
\label{tab:token_cost_all_parts}
\begin{adjustbox}{width=\columnwidth}
\begin{tabular}{clcc}
\toprule
Task Suite & Setting & Input Tokens $\downarrow$ & Output Tokens $\downarrow$\\
\midrule
\multirow{4}{*}{\shortstack{ParaView\\Visualization}}
& Claude-Code+Sonnet-4.5 (w/o skills) & \textbf{39.49M$\pm$6.62M} & 425.32K$\pm$55.52K \\
& Codex+GPT-5.2 (w/o skills) & 45.57M$\pm$9.47M & 396.60K$\pm$23.27K \\
& Claude-Code+Sonnet-4.5 (w/ skills) & 40.26M$\pm$4.37M & \textbf{101.04K$\pm$1.01K} \\
& Codex+GPT-5.2 (w/ skills) & 46.86M$\pm$3.78M & 329.25K$\pm$60.90K \\
\midrule
\multirow{4}{*}{\shortstack{Molecular\\Visualization}}
& Claude-Code+Sonnet-4.5 (w/o skills) & \textbf{5.07M$\pm$0.12M} & 81.73K$\pm$3.45K \\
& Codex+GPT-5.2 (w/o skills) & 8.63M$\pm$1.81M & 112.28K$\pm$17.22K \\
& Claude-Code+Sonnet-4.5 (w/ skills) & 7.64M$\pm$0.87M & \textbf{33.96K$\pm$1.90K} \\
& Codex+GPT-5.2 (w/ skills) & 11.95M$\pm$0.68M & 134.76K$\pm$8.83K \\

\midrule
\multirow{4}{*}{\shortstack{Bioimage\\Visualization}}
& Claude-Code+Sonnet-4.5 (w/o skills) & 8.60M$\pm$0.17M & 125.66K$\pm$2.43K \\
& Codex+GPT-5.2 (w/o skills) & 10.36M$\pm$3.76M & 104.83K$\pm$25.78K \\
& Claude-Code+Sonnet-4.5 (w/ skills) & \textbf{6.75M$\pm$0.73M} & \textbf{104.20K$\pm$9.73K} \\
& Codex+GPT-5.2 (w/ skills) & 17.41M$\pm$1.84M & 125.00K$\pm$8.89K \\
\midrule
\multirow{4}{*}{\shortstack{Topology\\Visualization}}
& Claude-Code+Sonnet-4.5 (w/o skills) & 17.26M$\pm$1.87M & 172.37K$\pm$18.51K \\
& Codex+GPT-5.2 (w/o skills) & 46.04M$\pm$4.48M & 193.58K$\pm$27.62K \\
& Claude-Code+Sonnet-4.5 (w/ skills) & \textbf{6.18M$\pm$1.59M} & 118.77K$\pm$30.97K \\
& Codex+GPT-5.2 (w/ skills) & 19.42M$\pm$3.56M & \textbf{112.51K$\pm$12.08K} \\

\midrule
\multirow{4}{*}{\shortstack{Object\\Identification}}
& Claude-Code+Sonnet-4.5 (w/o skills) & 16.83M$\pm$0.28M & 273.90K$\pm$22.71K \\
& Codex+GPT-5.2 (w/o skills) & 38.43M$\pm$3.55M & 344.27K$\pm$14.95K \\
& Claude-Code+Sonnet-4.5 (w/ skills) & \textbf{14.62M$\pm$0.91M} & \textbf{52.53K$\pm$0.92K} \\
& Codex+GPT-5.2 (w/ skills) & 31.65M$\pm$4.32M & 272.47K$\pm$24.59K \\
\bottomrule
\end{tabular}
\end{adjustbox}
\end{table}


\section{Experiments}

Prior work shows that general-purpose coding agents, specifically Codex and Claude Code, can outperform many domain-specific SciVis agents~\cite{ai2026scivisagentbench}. We evaluate \textit{SciVisAgentSkills} by comparing both agents with and without skills across all five task suites of SciVisAgentBench.

Each configuration is executed over three independent trials using the standardized evaluation pipeline of SciVisAgentBench, which combines multimodal LLM judges, image-based metrics, code validators, rule-based checks, and case-specific evaluators for outcome-based assessment. Table~\ref{tab:main_results} and Figure~\ref{fig:pass_k_combined} report the primary results using Claude-Opus-4.6 as the judge, which has been shown to align well with human SciVis expert assessments~\cite{ai2026scivisagentbench}. Completion rate measures whether runs terminate without execution errors, while pass metrics additionally require valid outputs (e.g., correctly saved, non-empty visualizations), providing a stricter measure of task success. Figure~\ref{fig:pass_k_combined} reports pass@${1,2,3}$ and pass$^{{1,2,3}}$ to characterize run-to-run consistency. For ParaView tasks, we additionally report image-based quality metrics (PSNR, SSIM, and LPIPS) in Table~\ref{tab:paraview_image_metrics}, using scaled aggregates to avoid overestimating partially completed results.

Overall, incorporating agent skills consistently improves performance across all task suites for the overall score, though the magnitude of improvement varies by task type and agent. The largest improvement is in Claude Code's topology visualization (about 60\%). More broadly, weaker baselines tend to benefit more from skills, highlighting the value of procedural guidance in challenging settings. Completion rate does not always follow the same trend as score: for example, Codex on object identification improves in overall score but drops from $92.59\pm9.80$ to $80.25\pm5.66$ in completion rate, indicating that skills can improve output quality while still introducing additional execution paths that sometimes fail.

We also analyze token usage in Table~\ref{tab:token_cost_all_parts}. For Claude Code, output tokens decrease consistently, and input tokens decrease for bioimage, topology, and object identification while increasing for molecular visualization. In contrast, Codex shows mixed trends, with increased token usage in bioimage and molecular visualization. Because the same skill content can reduce tokens in Claude Code but increase them in Codex, we interpret token cost as a property of the interaction among skill content, model behavior, and harness-level context management rather than as a property of skill verbosity alone. Overall, we do not observe a clear correlation between token usage and performance.

\section{Discussion}

{\bf Skills, CLI, and MCP.}
Skills, CLI access, and MCP-based tool calls are orthogonal design axes. A skill is a knowledge layer that describes how to use a CLI, a Python API, an MCP server, or a combination of these interfaces. Interactions through CLI are typically more flexible and token-efficient, and their performance further improves when augmented with agent skills that encode tool usage patterns. In contrast, protocols like MCP provide more structured and constrained interfaces, which can improve reliability but often at the cost of higher token usage and reduced generalization across tasks. Our experiments do not directly isolate CLI-MCP interaction paradigms. Rather, they demonstrate that procedural knowledge encoded as skills can improve agent performance regardless of the underlying interaction interface. A systematic comparison of CLI- and MCP-based interaction is beyond the scope of this work and has been investigated separately in recent SciVis agent studies~\cite{vonderhorst2026exploring}.

{\bf When do skills help, and when do they fail?}
Our four skills share a common recipe: fix the tool version, distill official documentation, reuse SciVis agent exemplars, and encode empirical fixes. Viewed more broadly, this recipe can be interpreted as a form of procedural knowledge distillation for scientific software. Rather than relying on agents to repeatedly rediscover environment assumptions, API conventions, and common failure modes during execution, agent skills externalize this knowledge into reusable artifacts that can be shared across tasks and users. This perspective may provide a useful framework for building skills beyond the specific tools studied here.

Across tasks, however, we see two recurring limits. First, gains may be limited when the base model already handles well-documented tools effectively; for example, Claude Code shows only modest improvement over VMD for molecular visualization with skills. Second, the interaction between skills and the underlying agent harness appears to be model-dependent. While Codex shows increased token usage on molecular and bioimage tasks, the same skills substantially reduce token usage for Claude Code. This suggests that token cost is not determined solely by skill verbosity. Instead, it likely depends on how the agent retrieves, caches, and reuses skill content during execution. Claude Code appears to benefit from more efficient use of skills, whereas Codex may repeatedly revisit skill content during planning and verification. Understanding these interactions remains an important direction for future work.

We also observe that skill benefits vary substantially across task suites. The largest improvements occur in topology visualization, object identification, and ParaView workflows, which require multi-step reasoning and involve APIs or workflow patterns that are less commonly represented in the training of foundation models. In contrast, gains in molecular visualization are smaller, particularly for Claude Code, suggesting that strong foundation models may already possess substantial knowledge of mature and widely documented tools such as VMD. These results indicate that agent skills are most valuable when procedural knowledge is specialized, fragmented across sources, or poorly represented in the model's training data.

{\bf Beyond skills: the role of the agent harness.}
Agent skills alone are not sufficient to support long-horizon workflows. Recent work on agent harness design emphasizes that reliable agents depend on the surrounding system that manages execution, context, and recovery~\cite{anthropic2025harness}. In particular, long-running agents require persistent state (e.g., progress tracking), structured interaction protocols, and mechanisms for incremental execution and rollback to maintain consistency across sessions~\cite{anthropic2026harness_design}. Recent work on AI VIS co-scientists further demonstrates that complex SciVis tasks benefit from specialized harness components, including orchestrators, domain-specific subagents, evaluation loops, memory systems, and tool integrations that coordinate end-to-end workflows~\cite{miao2026toward}. From this perspective, skills act as reusable procedural knowledge, while the harness provides the infrastructure for planning, memory, and control. Future work should therefore study skills and harnesses jointly, rather than in isolation.

{\bf Call for collaboration.}
We view this work as a starting point toward a broader ecosystem of agent skills for scientific data analysis and visualization. Beyond the performance improvements themselves, the consistent gains observed across ParaView, napari, VMD, and TTK suggest that agent skills can generalize across substantially different scientific software ecosystems rather than being tied to a single tool. We invite the community to contribute new skills, share evaluation cases, and explore how skill design interacts with agent harnesses in real-world, long-horizon workflows.

\section{Conclusions}

We present \textit{SciVisAgentSkills}, a set of domain-specific agent skills that enhance general-purpose coding agents for scientific data analysis and visualization. Experiments on SciVisAgentBench show that these skills improve mean performance across diverse tools and task types, while token usage depends on the agent harness and how skill context is managed. Our results suggest that improving long-horizon agent performance is not solely a matter of stronger models, but also of better procedural knowledge and system harness design. In particular, we believe that agent skills offer a low-cost and extensible entry point for improving practical agent performance in complex scientific settings.

\newpage
\acknowledgments{
This research was supported in part by the U.S.\ National Science Foundation through grants IIS-2101696, OAC-2104158, and IIS-2401144, and the U.S.\ Department of Energy through grant DE-SC0023145. 
This work was also performed under the auspices of the U.S. Department of Energy by Lawrence Livermore National Laboratory under Contract DE-AC52-07NA27344. The work is partially funded by DOE ECRP 51917/SCW1885. The manuscript is reviewed and released under LLNL-PROC-2019350.
}

\bibliographystyle{abbrv-doi-narrow}

\bibliography{refs}

@article{dhanoa2025agentic,
  title={Agentic Visualization: Extracting Agent-based Design Patterns from Visualization Systems},
  author={Dhanoa, Vaishali and Wolter, Anton and Le{\'o}n, Gabriela Molina and Schulz, Hans-J{\"o}rg and Elmqvist, Niklas},
  journal={IEEE Computer Graphics and Applications},
  volume = {45},
  number = {6},
  pages = {89--90},  
  year={2025},
  doi={10.1109/MCG.2025.3607741}
}

@inproceedings{liu2023agentbench,
  title={{AgentBench}: Evaluating {LLMs} as Agents},
  author={Liu, Xiao and Yu, Hao and Zhang, Hanchen and Xu, Yifan and Lei, Xuanyu and Lai, Hanyu and Gu, Yu and Ding, Hangliang and Men, Kaiwen and Yang, Kejuan and others},
  booktitle={Proceedings of International Conference on Learning Representations},
  year={2023},
  doi={10.48550/arXiv.2308.03688}
}

@misc{anthropic2024mcp,
  author       = {{Anthropic}},
  year          = {2024},
  title        = {Announcements: Introducing the Model Context Protocol},
  howpublished          = {\url{https://www.anthropic.com/news/model-context-protocol}}
}

@inproceedings{galimzyanov2025drawing,
  title={{Drawing Pandas}: A Benchmark for {LLMs} in Generating Plotting Code},
  author={Galimzyanov, Timur and Titov, Sergey and Golubev, Yaroslav and Bogomolov, Egor},
  booktitle={Proceedings of IEEE/ACM International Conference on Mining Software Repositories},
  pages={503--507},
  year={2025},
  doi={10.48550/arXiv.2412.02764}
}

@inproceedings{mialon2023gaia,
  title={{GAIA}: A Benchmark for General {AI} Assistants},
  author={Mialon, Gr{\'e}goire and Fourrier, Cl{\'e}mentine and Wolf, Thomas and LeCun, Yann and Scialom, Thomas},
  booktitle={Proceedings of International Conference on Learning Representations},
  year={2023},
  doi={10.48550/arXiv.2311.12983}
}

@article{chen2024viseval,
	author = {Chen, Nan and Zhang, Yuge and Xu, Jiahang and Ren, Kan and Yang, Yuqing},
	journal = {IEEE Transactions on Visualization and Computer Graphics},
	number = {1},
	pages = {1301--1311},
	title = {{VisEval}: A Benchmark for Data Visualization in the Era of Large Language Models},
	volume = {31},
	year = {2025},
    doi = {10.1109/TVCG.2024.3456320}
}

@article{yao2024tau,
  title={$\tau$-bench: A Benchmark for Tool-Agent-User Interaction in Real-World Domains},
  author={Shunyu Yao and Noah Shinn and Pedram Razavi and Karthik Narasimhan},
  journal={arXiv preprint arXiv:2406.12045},
  year={2024},
  doi={10.48550/arXiv.2406.12045}
}

@inproceedings{liu2025paraview,
    author = {Shusen Liu and Haichao Miao and Peer-Timo Bremer},
    booktitle = {Proceedings of IEEE VIS Conference (Short Papers)},
    pages = {61--65},
    title = {{ParaView-MCP}: An Autonomous Visualization Agent with Direct Tool Use},
    year = {2025},
    doi={10.48550/arXiv.2505.07064}
}

@inproceedings{dibia2023lida,
  title={{LIDA}: A Tool for Automatic Generation of Grammar-Agnostic Visualizations and Infographics using Large Language Models},
  author={Dibia, Victor},
  booktitle={Proceedings of Annual Meeting of the Association for Computational Linguistics: System Demonstrations},
  pages={113--126},
  year={2023},
  doi={10.18653/v1/2023.acl-demo.11}
}

@article{ai2025nli4volvis,
 author = "Kuangshi Ai and Kaiyuan Tang and Chaoli Wang",
 title = "{NLI4VolVis}: Natural Language Interaction for Volume Visualization via Multi-{LLM} Agents and Editable {3D Gaussian} Splatting",
 journal = "IEEE Transactions on Visualization and Computer Graphics",
 volume = "32",
 number = "1",
 year = "2026",
 pages = "46--56",
 doi = "10.1109/TVCG.2025.3633888"
}

@article{liu2024ava,
  title={{AVA}: Towards Autonomous Visualization Agents through Visual Perception-Driven Decision-Making},
  author={Liu, Shusen and Miao, Haichao and Li, Zhimin and Olson, Matthew and Pascucci, Valerio and Bremer, Peer-Timo},
  journal={Computer Graphics Forum},
  volume={43},
  number={3},
  pages={e15093},
  year={2024},
  doi={10.1111/cgf.15093}
}

@inproceedings{peterka2025chatvis,
author = {Peterka, Tom and Mallick, Tanwi and Yildiz, Orcun and Lenz, David and Quammen, Cory and Geveci, Berk},
booktitle = {Proceedings of IEEE Workshop on Large Data Analysis and Visualization},
title = {{ChatVis}: Large Language Model Agent for Generating Scientific Visualizations},
year = {2025},
pages = {22--32},
doi = {10.1109/LDAV68558.2025.00007},
}

@article{wang2025intuitf,
  title={{IntuiTF}: {MLLM}-Guided Transfer Function Optimization for Direct Volume Rendering},
  author={Wang, Yiyao and Pan, Bo and Wang, Ke and Liu, Han and Mao, Jinyuan and Liu, Yuxin and Zhu, Minfeng and Zhang, Bo and Chen, Weifeng and Huang, Xiuqi and others},
  journal={arXiv preprint arXiv:2506.18407},
  year={2025},
  doi={10.48550/arXiv.2506.18407}
}

@article{jia2025voice,
  author={Jia, Donggang and Irger, Alexandra and Besançon, Lonni and Strnad, Ondřej and Luo, Deng and Björklund, Johanna and Kouyoumdjian, Alexandre and Ynnerman, Anders and Viola, Ivan},
  journal={IEEE Transactions on Visualization and Computer Graphics}, 
  title={{VOICE}: Visual Oracle for Interaction, Conversation, and Explanation}, 
  year={2025},
  volume={31},
  number={10},
  pages={8828--8845},
  doi={10.1109/TVCG.2025.3579956}
}

@article{chen2025coda,
  title={{CoDA}: Agentic Systems for Collaborative Data Visualization},
  author={Chen, Zichen and Chen, Jiefeng and Arik, Sercan {\"O} and Sra, Misha and Pfister, Tomas and Yoon, Jinsung},
  journal={arXiv preprint arXiv:2510.03194},
  year={2025},
  doi={10.48550/arXiv.2510.03194}
}

@inproceedings{tam2025infera,
  title={{InferA}: A Smart Assistant for Cosmological Ensemble Data},
  author={Tam, Justin Z. and Grosset, Pascal and Banesh, Divya and Ramachandra, Nesar and Turton, Terece L. and Ahrens, James P.},
  booktitle={Proceedings of ACM/IEEE SC Workshops},
  pages={20--28},
  year={2025},
  doi={10.1145/3731599.3767342}
}

@article{biswas2025vizgenie,
	author = {Ayan Biswas and Terece L. Turton and Nishath Rajiv Ranasinghe and Shawn Jones and Bradley Love and William Jones and Aric Hagberg and Han-Wei Shen and Nathan DeBardeleben and Earl Lawrence},
	journal = {IEEE Transactions on Visualization and Computer Graphics},
	pages = {1021--1031},
	number = {1},
	title = {{VizGenie}: Toward Self-Refining, Domain-Aware Workflows for Next-Generation Scientific Visualization},
	volume = {32},
	year = {2026},
    doi = {10.1109/TVCG.2025.3634655}
}

@article{tang2025texgs,
 author = "Kaiyuan Tang and Kuangshi Ai and Jun Han and Chaoli Wang",
 title = "{TexGS-VolVis}: Expressive Scene Editing for Volume Visualization via Textured {Gaussian} Splatting",
 journal = "IEEE Transactions on Visualization and Computer Graphics",
 volume = "32",
 number = "1",
 year = "2026",
 pages = "933--943" ,
 doi = "10.1109/TVCG.2025.3634643"
}

@inproceedings{Ai-GenAI25,
 author = {Kuangshi Ai and Haichao Miao and Zhimin Li and Chaoli Wang and Shusen Liu},
 title = {An Evaluation-Centric Paradigm for Scientific Visualization Agents},
 booktitle = {Proceedings of IEEE Workshop on GenAI, Agents, and the Future of VIS},
 year = {2025},
 doi = {10.48550/arXiv.2509.15160}
}

@incollection{Ahrens2005ParaView,
  title={{ParaView}: An End-User Tool for Large-Data Visualization},
  author={James P. Ahrens and Berk Geveci and C. Charles Law},
  booktitle={The Visualization Handbook},
  editor={Charles D. Hansen and Christopher R. Johnson},
  chapter={36},
  pages={717--731},
  publisher={Academic Press},
  year={2004},
  doi={10.1016/B978-012387582-2/50038-1}
}

@article{VMD1996,
  author={William Humphrey and Andrew Dalke and Klaus Schulten},
  title={{VMD}: Visual Molecular Dynamics},
  journal={Journal of Molecular Graphics},
  year={1996},
  volume={14},
  pages={33--38},
  doi={10.1016/0263-7855(96)00018-5}
}

@misc{napari2019,
  author       = {Sofroniew, Nicholas and Lambert, Talley and Bokota, Grzegorz and Nunez-Iglesias, Juan and Sobolewski, Peter and Sweet, Andrew and Gaifas, Lorenzo and Evans, Kira and Burt, Alister and Doncila Pop, Draga and others},
  title        = {napari: A Multi-Dimensional Image Viewer for {Python}},
  year         = {2025},
  publisher    = {Zenodo},
  doi          = {10.5281/zenodo.3555620}
}

@misc{egtai2025gmxvmdmcp,
  author       = {{EgT}},
  title        = {{GMX-VMD-MCP}: {MCP} Service for {GROMACS} and {VMD} Molecular Dynamics Simulations and Visualization},
  year         = {2025},
  howpublished = {\url{https://github.com/egtai/gmx-vmd-mcp}}
}

@misc{llnl2025bioimageagent,
  author       = {Haichao Miao and Shusen Liu},
  title        = {{BioImage-Agent}},
  year         = {2025},
  howpublished = {\url{https://github.com/LLNL/bioimage-agent}}
}

@misc{anthropic2025claudecode,
  author       = {{Anthropic}},
  title        = {{Claude Code}: An Agentic Coding Tool},
  year         = {2025},
  howpublished          = {\url{https://github.com/anthropics/claude-code}}
}

@misc{openai2025codex,
  author       = {{OpenAI}},
  title        = {{OpenAI Codex}: Lightweight Coding Agent that Runs in Your Terminal},
  year         = {2025},
  howpublished = {\url{https://github.com/openai/codex}}
}

@misc{google2025geminicli,
  author       = {{Google}},
  title        = {{Gemini CLI}: An Open-Source AI Agent that Brings the Power of Gemini Directly into Your Terminal},
  year         = {2025},
  howpublished = {\url{https://github.com/google-gemini/gemini-cli}}
}

@article{do2026svlat,
  title={{SVLAT}: Scientific Visualization Literacy Assessment Test},
  author={Do, Patrick Phuoc and Tang, Kaiyuan and Ai, Kuangshi and Wang, Chaoli},
  journal={arXiv preprint arXiv:2603.19000},
  year={2026},
  doi={10.48550/arXiv.2603.19000}
}

@article{gorski2026topopilot,
  title={{TopoPilot}: Reliable Conversational Workflow Automation for Topological Data Analysis and Visualization},
  author={Nathaniel Gorski and Shusen Liu and Bei Wang},
  journal={arXiv preprint arXiv:2603.25063},
  year={2026},
  doi={10.48550/arXiv.2603.25063}
}

@article{li2026skillsbench,
  title={{SkillsBench}: Benchmarking How Well Agent Skills Work Across Diverse Tasks},
  author={Li, Xiangyi and Chen, Wenbo and Liu, Yimin and Zheng, Shenghan and Chen, Xiaokun and He, Yifeng and Li, Yubo and You, Bingran and Shen, Haotian and Sun, Jiankai and others},
  journal={arXiv preprint arXiv:2602.12670},
  year={2026},
  doi={10.48550/arXiv.2602.12670}
}

@article{jiang2026sok,
  title={{SoK}: Agentic Skills--Beyond Tool Use in {LLM} Agents},
  author={Jiang, Yanna and Li, Delong and Deng, Haiyu and Ma, Baihe and Wang, Xu and Wang, Qin and Yu, Guangsheng},
  journal={arXiv preprint arXiv:2602.20867},
  year={2026},
  doi={10.48550/arXiv.2602.20867}
}

@article{sun2026sasav,
  title={{SASAV}: Self-Directed Agent for Scientific Analysis and Visualization},
  author={Sun, Jianxin and Lenz, David and Peterka, Tom and Yu, Hongfeng},
  journal={arXiv preprint arXiv:2604.03406},
  year={2026},
  doi={10.48550/arXiv.2604.03406}
}

@article{ai2026scivisagentbench,
  title={{SciVisAgentBench}: A Benchmark for Evaluating Scientific Data Analysis and Visualization Agents},
  author={Ai, Kuangshi and Miao, Haichao and Tang, Kaiyuan and Gorski, Nathaniel and Sun, Jianxin and Liu, Guoxi and Ing{\'o}lfsson, Helgi I and Lenz, David and Guo, Hanqi and Yu, Hongfeng and others},
  journal={arXiv preprint arXiv:2603.29139},
  year={2026},
  doi={10.48550/arXiv.2603.29139}
}

@article{xu2026agent,
  title={Agent Skills for Large Language Models: Architecture, Acquisition, Security, and the Path Forward},
  author={Xu, Renjun and Yan, Yang},
  journal={arXiv preprint arXiv:2602.12430},
  year={2026},
  doi={10.48550/arXiv.2602.12430}
}

@article{ling2026agent,
  title={Agent Skills: A Data-Driven Analysis of Claude Skills for Extending Large Language Model Functionality},
  author={Ling, George and Zhong, Shanshan and Huang, Richard},
  journal={arXiv preprint arXiv:2602.08004},
  year={2026},
  doi={10.48550/arXiv.2602.08004}
}

@misc{anthropic2025agentskillsblog,
  author       = {{Anthropic}},
  title        = {Equipping Agents for the Real World with Agent Skills},
  year         = {2025},
  howpublished = {\url{https://claude.com/blog/equipping-agents-for-the-real-world-with-agent-skills}}
}

@article{ttk2018,
  author={Tierny, Julien and Favelier, Guillaume and Levine, Joshua A. and Gueunet, Charles and Michaux, Michael},
  journal={IEEE Transactions on Visualization and Computer Graphics}, 
  title={The Topology ToolKit}, 
  year={2018},
  volume={24},
  number={1},
  pages={832--842},
  doi={10.1109/TVCG.2017.2743938}
}

@misc{anthropic2025harness,
  title        = {Effective Harnesses for Long-Running Agents},
  author       = {{Anthropic}},
  year         = {2025},
  howpublished = {\url{https://www.anthropic.com/engineering/effective-harnesses-for-long-running-agents}}
}

@misc{anthropic2026harness_design,
  title = {Harness Design for Long-Running Application Development},
  author = {{Anthropic}},
  year = {2026},
  howpublished = {\url{https://www.anthropic.com/engineering/harness-design-long-running-apps}}
}

@article{vonderhorst2026exploring,
  title={Exploring Interaction Paradigms for {LLM} Agents in Scientific Visualization},
  author={Vonderhorst, Jackson and Ai, Kuangshi and Miao, Haichao and Liu, Shusen and Wang, Chaoli},
  journal={arXiv preprint arXiv:2604.27996},
  year={2026},
  doi={10.48550/arXiv.2604.27996}
}

@inproceedings{mathai2026nl2scivis,
booktitle = {Proceedings of Eurographics Conference on Visualization (Short Papers)},
title = {{NL2SciVis}: A Benchmark for Natural Language to Scientific Visualization},
author = {Mathai, Manish and Han, Mengjiao and Knowles, Janet and Mateevitsi, Victor A. and Rizzi, Silvio and Childs, Hank},
year = {2026},
doi = {10.2312/evs.20261017}
}

@article{miao2026toward,
  title={Toward {AI VIS} Co-Scientists: A General and End-to-End Agent Harness for Solving Complex Data Visualization Tasks},
  author={Miao, Haichao and Li, Zhimin and Ai, Kuangshi and Tang, Kaiyuan and Wang, Chaoli and Bremer, Peer-Timo and Liu, Shusen},
  journal={arXiv preprint arXiv:2605.21825},
  year={2026},
  doi={10.48550/arXiv.2605.21825}
}
\end{document}